\documentclass{article}
\usepackage{spconf,amsmath,graphicx}
\usepackage[linesnumbered,ruled]{algorithm2e}
\usepackage{kotex}
\usepackage{multirow}
\usepackage{comment}
\usepackage{amsmath}
\usepackage{mathtools}
\DeclareMathOperator*{\argmax}{arg\,max}

\newcommand{\algofn}{\texttt}

\title{Label Propagation Adaptive Resonance Theory\\for Semi-supervised Continuous Learning}

\name{\begin{tabular}{c}Taehyeong Kim$^{1,2}$ \qquad Injune Hwang$^{1}$ \qquad Gi-Cheon Kang$^{1}$ \qquad Won-Seok Choi$^{1}$ \\ Hyunseo Kim$^{1}$ \qquad Byoung-Tak Zhang$^{1}$\end{tabular}
\thanks{This work was partly supported by the Korea government (2019-0-01367-BabyMind, 2015-0-00310-SW.StarLab, 2017-0-01772-VTT, 2018-0-00622-RMI, P0006720-GENKO).}}

\address{$^{1}$Seoul National University, Seoul, Republic of Korea \\
$^{2}$AI Lab, CTO Division, LG Electronics, Seoul, Republic of Korea \\
\small{\{thkim, ijhwang, gckang, wchoi, hskim, btzhang\}@bi.snu.ac.kr}}

\begin{document}
%
\maketitle
\begin{abstract}
Semi-supervised learning and continuous learning are fundamental paradigms for human-level intelligence.
To deal with real-world problems where labels are rarely given and the opportunity to access the same data is limited, it is necessary to apply these two paradigms in a joined fashion.
In this paper, we propose \textit{Label Propagation Adaptive Resonance Theory} (LPART) for semi-supervised continuous learning.
LPART uses an online label propagation mechanism to perform classification and gradually improves its accuracy as the observed data accumulates.
We evaluated the proposed model on visual (MNIST, SVHN, CIFAR-10) and audio (NSynth) datasets by adjusting the ratio of the labeled and unlabeled data.
The accuracies are much higher when both labeled and unlabeled data are used, demonstrating the significant advantage of LPART in environments where the data labels are scarce.
\end{abstract}


\section{Introduction}
\label{sec:intro}
Over the last few years, the deep neural network models have shown remarkable progresses especially in visual object recognition, speech recognition and autonomous robot control.
However, using deep learning has a major practical shortcoming that it requires time and labor not only to collect a massive amount of data but also to label them.
In this aspect, the fields of semi-supervised learning, continuous learning, transfer learning and meta-learning have been in the spotlight.

The semi-supervised learning paradigm \cite{zhu2009introduction} tackles the problem in which the unlabeled data is abundant but the labeled data is extremely limited. 
The continuous learning paradigm \cite{mccloskey1989catastrophic} aims to learn without catastrophic forgetting of the former knowledge from sequential data, allowing the model to adapt to the ever-changing environments.
Although each paradigm is promising on its own, we argue that both paradigms should be applied in a joined fashion to deal with many real-world problems, where labeled data is rarely given and the data once learned is hard to access again.

In this study, we propose \textit{Label Propagation Adaptive Resonance Theory} (LPART) for semi-supervised continuous learning.
Adaptive Resonance Theory (ART) is a solution to continuous learning inspired by brain information processing mechanisms \cite{grossberg1987competitive}, and various label propagation methods have been studied for semi-supervised learning \cite{iscen2019label, douze2018low}.
However, these label propagation methods are not suitable for the environment with limited opportunity to access the same data because they require the data repeatedly to be learned.

Therefore, we propose an online label propagation method for continuous learning in the ART network.
Specifically, we use a two-fold learning process: (1) feature extraction using variational autoencoders (VAE) \cite{kingma2013auto} and (2) clustering of the extracted features and inference of the classes using LPART.
First, we train VAE in a weakly-supervised manner by using the \textit{pair loss}.
Then, LPART takes the features as input and learns to infer the classes of unlabeled nodes by leveraging the label propagation method.
If the amount of labeled data is not enough, the inference of the unlabeled node could be inaccurate.
Therefore, we use the metrics to measure the uncertainties so that LPART could defer its classification decisions for the nodes with high uncertainty.
We experimentally confirmed that our proposed model is able to learn continuously without catastrophic forgetting, even with the rarely labeled data.

\section{Adaptive Resonance Theory}
\label{sec:preliminary}

ART is a self-organizing neural network inspired by the brain information processing mechanisms.
ART uses the interaction of `top-down' expectation and `bottom-up' sensory information to learn adaptively, using resonance.
There are two general principles in ART: (1) The knowledge is strengthened when the sensation is strong enough and the expectation matches well with the sensation, and (2) if there is no expectation that matches the sensation, new knowledge is learned.
In terms of being conservative while learning new, the ART system can be a solution for the continuous learning.

Various ART networks such as Fuzzy ART \cite{carpenter1991fuzzy}, ART- MAP \cite{carpenter1991artmap}, and Fuzzy ARTMAP (FAM) \cite{carpenter1992fuzzy} have been studied.
Fuzzy ART can process real-valued data by using fuzzy set theory.
ARTMAP uses supervised learning and classification system that is built up from a pair of ART modules.
FAM integrates the advantages of both Fuzzy ART and ARTMAP.

There are also other ART networks for learning in a semi-supervised manner, such as semi-supervised Bayesian ART- MAP (SSBA) \cite{TangSemiART2009} and semi-supervised Fuzzy ARTMAP (ssFAM) \cite{beART}.
SSBA employs EM algorithm based on Bayesian ARTMAP (BA) \cite{vigdor2007bayesian} to adjust its parameters, which realizes the soft assignment of training samples instead of the winner-take-all strategy.
ssFAM relies on FAM but adopts a tunable network parameter called \textit{category prediction error tolerance}, which achieves semi-supervised learning.

The present study differs from previous studies in that it applies an online label propagation mechanism for semi-supervised continuous learning.
In addition, the label propagation mechanism can be applied to various kinds of ART networks, which allows the extension from LPART.
Also, the uncertainty measurement methods proposed in this study can be used to filter reliable classification results.

\section{Methods}
\label{sec:methodology}

\subsection{Feature Extraction with VAE}
\label{sec:vae}

It is necessary to extract features that are easy to cluster for ART to classify high-dimensional data properly.
VAE, a deep learning-based unsupervised learning method, is widely used to extract useful features from data \cite{hou2017deep, pu2016variational}.
However, basic VAE does not have any explicit constraints to improve clustering.
In this regard, the study of learning representations using \textit{oracle triplets} provides the insights needed for this study \cite{karaletsos2015bayesian}.
We also repurposed the VAE architecture for semi-supervised continuous learning.

In this study, we use a simplified triplet-based VAE to extract features.
It uses only some dimensions $d$ in the latent space for VAE encoder to produce class-embedded representation $\mu_d$.
Additional \textit{pair loss} is introduced, which depends on whether or not the class of the previous sample and the current sample are the same.
The \textit{pair loss} between previous and current class-embedded representations is defined using the L2 distance as a similarity measure (Equation \ref{eq:pair_loss}).
\begin{equation}
\label{eq:pair_loss}
\mathcal{L}_{pair}=\begin{cases}
||\mu_{d,prev}-\mu_{d,curr}||_2, \quad y_{prev}=y_{curr} \\
-||\mu_{d,prev}-\mu_{d,curr}||_2, \thinspace otherwise
\end{cases}
\end{equation}
Here, $y$ denotes a label of the input sample.
We optimize parameters by maximizing the ELBO (evidence lower bound) \cite{kingma2013auto} and minimizing the \textit{pair loss}. With a scaling factor $\lambda$, the total loss to be minimized is as shown in Equation \ref{eq:losses}.
\begin{equation}
\label{eq:losses}
\mathcal{L} = -ELBO + \lambda\mathcal{L}_{pair} \\
\end{equation}


\subsection{The LPART Algorithm}
\label{sec:lpart}

When an input data $x_i$ is given, we encode it using the VAE previously described to get the 0-to-1 normalized class-embedded representation as $r_i$.
As in Fuzzy ART, we also compute the complement coding $I_i$ of $r_i$.
For a node $j$ with a weight vector $w_j$, the choice function $T_j$ and the match function $V_j$ of $I_i$ are defined as:
\begin{equation}
\label{eq:activation}
T_j(I_i) = \frac{\left \| I_i \wedge w_j \right \|_1}{\alpha + \left \| w_j \right \|_1}, \qquad V_j(I_i)=\frac{\left \| I_i \wedge w_j \right \|_1}{\left \| I_i \right \|_1}
\end{equation}
where $\wedge$ is the element-wise minimum operator, $\alpha > 0$ is the choice parameter and $\| \cdot \|_1$ denotes the L1 norm of a vector.

If the value of $V_j(I_i)$ is greater than a vigilance parameter $\rho$, we say that the node $j$ has matched $x_i$ or been \textit{activated}.
Among all of the activated nodes, a \textit{winner} $J$ with the highest value of $T_j$ is selected.
It can be seen as the best-fit node for the input, and we update its weight vector considering $I_i$ with a learning rate $\beta$ between 0 and 1 as shown in Equation \ref{eq:update}.
\begin{equation}
\label{eq:update}
w^{new}_J=\beta (I_i \wedge w^{old}_J) + (1-\beta)w^{old}_J
\end{equation}

On the other hand, if no node matches the input, a new node is created with an initial parameter set as $I_i$.
This newborn node can grow larger throughout the subsequent iterations.
The creation of a node is more frequent with a larger value of vigilance parameter $\rho$.
By manipulating the value of $\rho$, we can balance the rigidity of the node.
If it gets too small, one node covers up too many inputs, making the consistency of the node vague.
Therefore, we use sufficiently large value for the vigilance parameter.

Another crucial part of LPART, the label propagation mechanism, will be explained in the following section.
The overall LPART algorithm is described in Algorithm \ref{alg:lpart}.

\begin{algorithm}
\label{alg:lpart}
\SetAlgoVlined
\SetKw{KwWithin}{in}
    \For(\scriptsize{\tcp*[f]{$y_i$ can be absent}}\normalsize{}){$x_i, y_i$ \KwWithin dataset}{
        $r_i \leftarrow \algofn{Encode}(x_i) $ \\
        $I_i \leftarrow [r_i, \overrightarrow{1} - r_i] $
       \scriptsize{\tcp*[f]{concatenation}}\normalsize{} \\
        $A \leftarrow \{\}$\\
        \For(\scriptsize{\tcp*[f]{\textit{N} is the number of nodes}}\normalsize{}){$j$ \KwWithin $1, \ldots, N$ } {
            $T_j \leftarrow \|I_i \wedge w_j\|_1 \; / \; (\alpha + \|w_j\|_1)$ \\
            $V_j \leftarrow \|I_i \wedge w_j\|_1 \; / \; \|I_i\|_1$ \\
            \If{$V_j \geq \rho$}{
                \If{$y_i$ is given}{
                    $q_j(y_i) \leftarrow q_j(y_i) + 1$
                }
                $A \leftarrow A \cup \{j\}$
            }
        }

        \eIf{A is not empty}{
            $\algofn{LabelPropagate}(A)$ \scriptsize{\tcp*[f]{if |A| > 1}}\normalsize{} \\
            $J \leftarrow \argmax_{j \in A}(T_j) $ \\
            $w_J \leftarrow \beta (I_i \wedge w_J) + (1 - \beta) w_J$
        }{
            $\algofn{CreateNode}(x_i, y_i)$ \scriptsize{\tcp*[f]{$q_n(y_i) \leftarrow 1$}}\normalsize{}
        }
    }
    \caption{The LPART algorithm.}
\end{algorithm}

\subsection{Label Propagation Mechanism}
Label propagation, a mechanism for semi-supervised learning, is a method of inferring a class of unlabeled cluster with the help of labeled ones \cite{Zhu02learningfrom}.
It assumes that the clusters close to each other in the feature space tend to belong to similar classes.
In LPART, label propagation is triggered when an input data activates two or more nodes.
The co-activated nodes can be considered to be located in the vicinity of each other in the feature space, which in turn implies a high relevance between them.
It is natural, therefore, to estimate the label of a label-absent node\textemdash a node that does not contain any input with a label\textemdash using the labels of co-activated nodes.
The numerical value of the label itself is meaningless, so we use a distribution over all labels instead of a single value.
We call this a \textit{label density function} and denote by $q$, where $q_j(y)$ roughly means how probable a node $j$ will be in class $y$.
When a new node $n$ is created, $q_n$ is initialized to the zero vector.

Once a labeled sample is added to a node, the density of its label increases by one.
For a label-absent node $k$, label density function is updated by averaging those of co-activated nodes:
\begin{equation}
\label{eq:prob_with_soft_label2}
\begin{aligned}
q_k^{new}(y)
= \left( \delta \times
\frac{\sum_{j \in A-\left\{k\right\}}q_j^{old}(y)}
{\sum_{y'}\sum_{j \in A-\left\{k\right\}}q_j^{old}(y')} \right. \\
+ \left. (1-\delta) \times
\frac{q_k^{old}(y)}{\sum_{y'}q_k^{old}(y')} \right)
\times \frac{1}{C}
\end{aligned}
\end{equation}
where $\delta$ is a propagation rate.
The reason why the sum of $q_k$ over all labels is less than one is to indicate that it is still not certain which class this node belongs to.
$C > 1$ can be interpreted as a kind of uncertainty parameter, which will be further discussed in Section \ref{sec:uncertainty}.

Finally, the probability distribution of labels for each node is easily obtained by normalizing the label density function:
\begin{equation}
\label{eq:prob_with_soft_label1}
p_j(y) = \frac{q_{j}(y)}{\sum_{y'}q_{j}(y')}
\end{equation}
and we can infer a class of input data by finding a winner node and then selecting a label with the highest probability.

\subsection{Measurement of Uncertainty}
\label{sec:uncertainty}

We use two different metrics to measure the uncertainty of the classification results.
The first uncertainty, $ u ^ 1 (x_ {i}) $, is measured by the entropy of the classification probability \cite{klir2013uncertainty}, as shown in Equation \ref{eq:entropy}.
This method measures how evenly distributed the categories of labeled data learned by each node, which can be used to identify high-impurity nodes.
\begin{equation}
\label{eq:entropy}
u^1(x_{i}) = -\sum_{y} p_{J(x_i)}(y)\log p_{J(x_i)}(y)
\end{equation}
$J(x_i)$ is a winning node with given input $x_i$ and $ p_{J(x_i)}(y) $ is the probability that this node belongs to class $y$.

The second uncertainty $ u ^ 2 (x_ {i}) $ is based on the number of labeled input data in each node as shown in Equation \ref{eq:count}.
It allows them to filter out highly unreliable recognition results from the datasets with insufficient number of labels.
\begin{equation}
\label{eq:count}
u^2(x_{i}) = 1 - tanh(k \cdot \Sigma_{y} q_{J(x_i)}(y))
\end{equation}
Here, $k$ is a constant for sensitivity.
The combination of these two uncertainties comes in handy for real-world problems; for example, we can withhold judgments on samples with high uncertainties during continuous learning.

\section{Experiments}
\label{sec:experiments}
\subsection{Dataset}
\label{sec:dataset}
\begin{figure}[htb]
\begin{minipage}[b]{0.24\linewidth}
  \centering
  \centerline{\includegraphics[width=2.0cm]{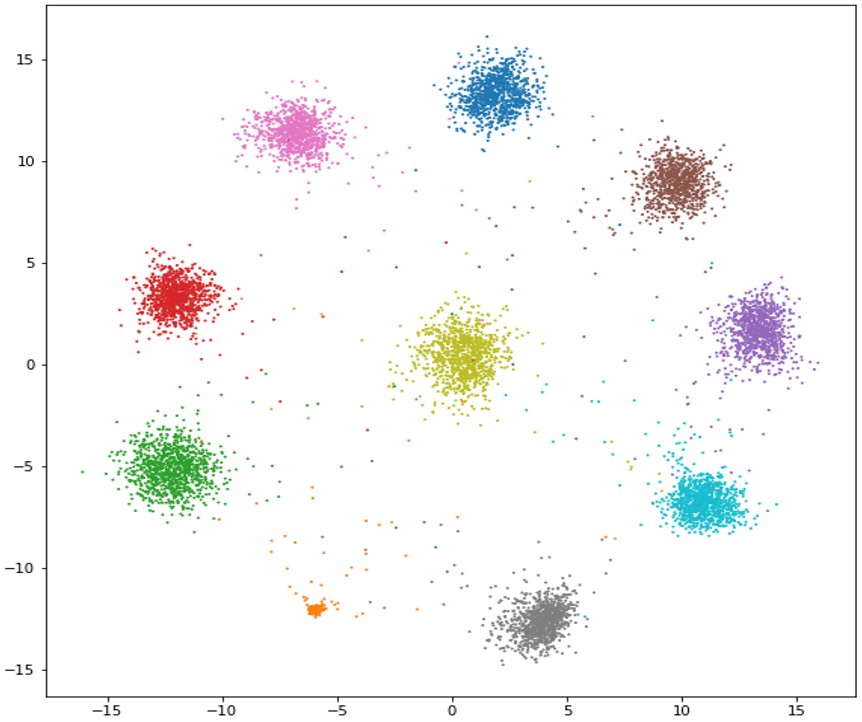}}
  \centerline{(a) MNIST}\medskip
\end{minipage}
\hfill
\begin{minipage}[b]{.24\linewidth}
  \centering
  \centerline{\includegraphics[width=2.0cm]{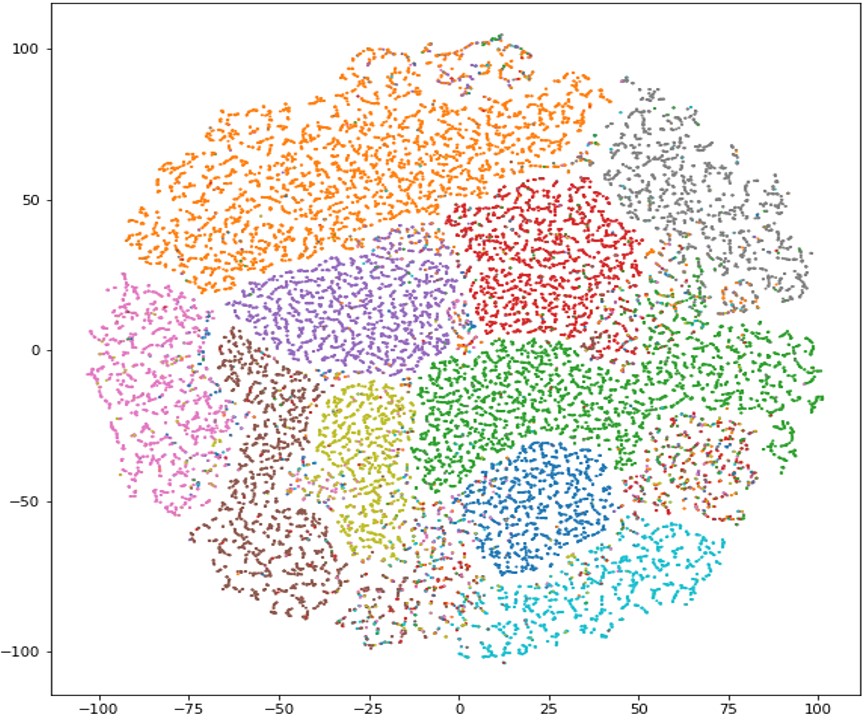}}
  \centerline{(b) SVHN}\medskip
\end{minipage}
\hfill
\begin{minipage}[b]{.24\linewidth}
  \centering
  \centerline{\includegraphics[width=2.0cm]{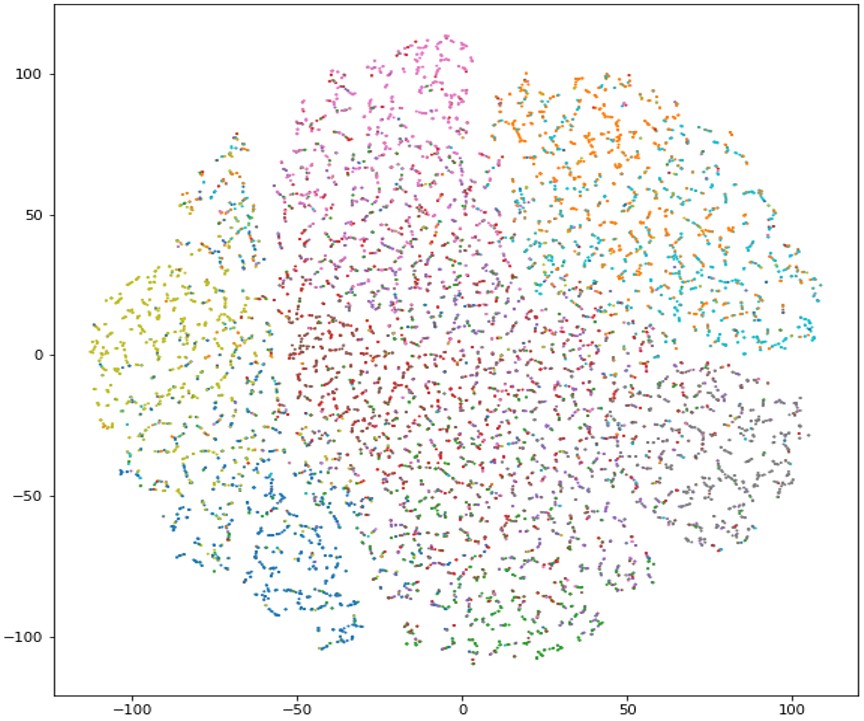}}
  \centerline{(c) CIFAR-10}\medskip
\end{minipage}
\hfill
\begin{minipage}[b]{.24\linewidth}
  \centering
  \centerline{\includegraphics[width=2.0cm]{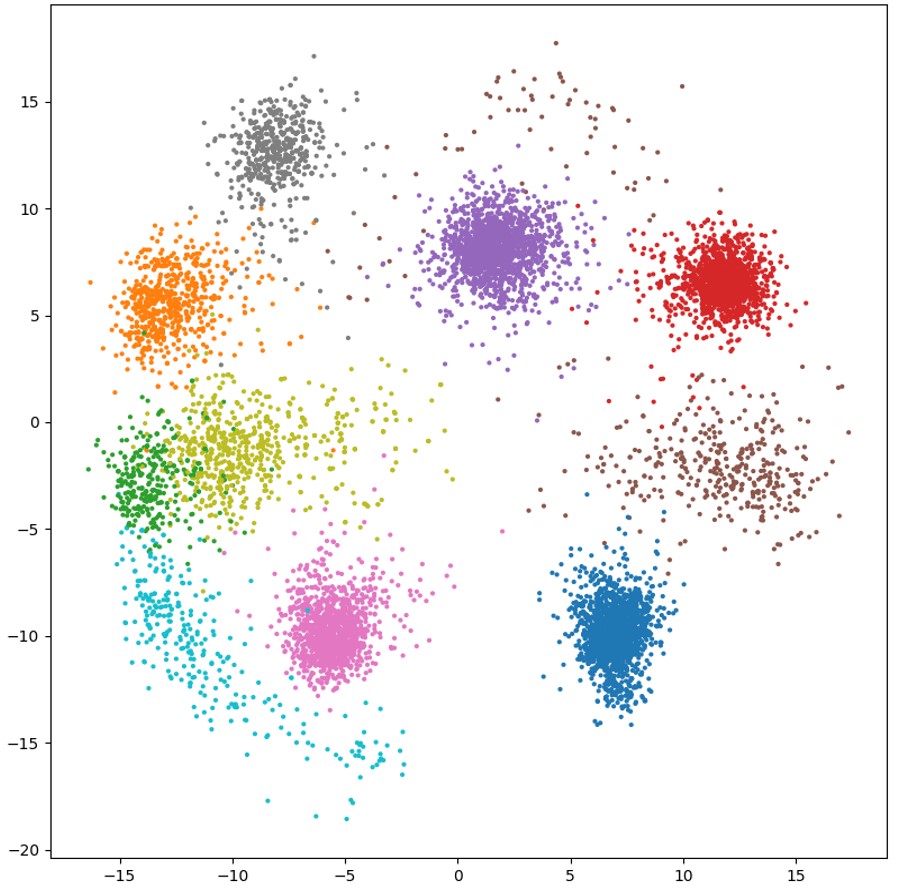}}
  \centerline{(d) NSynth}\medskip
\end{minipage}
\vspace{-11px}
\caption{The visualization of extracted features. For plotting (b) and (c), t-SNE \cite{maaten2008visualizing} was used to reduce dimensions.}
\label{fig:feature}
\vspace{-9px}
\end{figure}

We investigated our proposed model using MNIST \cite{lecun1998gradient}, SVHN \cite{netzer2011reading}, CIFAR-10 \cite{krizhevsky2009learning}, and NSynth \cite{engel2017neural} datasets.
The MNIST and SVHN datasets consist of the digit images from 0 to 9.
The CIFAR-10 dataset contains 60,000 color images in 10 classes such as airplanes, cars, birds and cats.
The NSynth dataset contains 305,979 4-second audio recordings from 1,006 instruments.
Each recording is labeled with one of the 11 high-level groups such as bass, keyboard, synth lead, and vocal.
Note that since we only used 12,678 validation split as training data (with 4,096 original test split), the synth lead included only in the training split was omitted.
All audio data was converted to spectrogram images for feature extraction.
The features extracted using the VAE are shown in Figure \ref{fig:feature}.

\begin{table*}
    \caption{Classification accuracy of our model (LPART) compared to FAM trained for a single epoch with various probabilities of the labeled data. The mean and standard deviation are drawn from 30 trials for each experiment. (unit : \%)}
    \vspace{-3px}
    \begin{center}
    \centerline{
        \begin{tabular}{c c c c c c c c c}
        \hline
        Dataset rate & \multicolumn{2}{c}{MNIST ($\rho=0.99$)} & \multicolumn{2}{c}{SVHN ($\rho=0.98$)} & \multicolumn{2}{c}{CIFAR-10 ($\rho=0.95$)} & \multicolumn{2}{c}{NSynth ($\rho=0.95$)}\\
        \cline{2-9}
        (labeled, unlabeled) & FAM & LPART & FAM & LPART & FAM & LPART & FAM & LPART \\
        \hline\hline
        0.1\%, not used & 46.2$\pm$3.0 & 46.5$\pm$3.4 & 48.1$\pm$2.8 & 48.8$\pm$2.9 & 33.0$\pm$2.5 & 30.9$\pm$2.9 & 46.8$\pm$9.3 & 48.3$\pm$10.9 \\
        0.1\%, 99.9\% & - & \textbf{94.2}$\pm$1.0 & - & \textbf{73.7}$\pm$1.3 & - & \textbf{39.5}$\pm$2.0 & - & \textbf{63.1}$\pm$11.7 \\
        0.5\%, not used & 71.8$\pm$1.9 & 70.9$\pm$1.7 & 60.5$\pm$2.0 & 59.4$\pm$1.9 & 38.8$\pm$1.2 & 37.1$\pm$1.5 & 67.2$\pm$4.3 & 68.3$\pm$4.5 \\
        0.5\%, 99.5\% & - & \textbf{95.0}$\pm$0.3 & - & \textbf{74.5}$\pm$0.4 & - & \textbf{42.4}$\pm$0.9 & - & \textbf{85.2}$\pm$2.1 \\
        1.0\%, not used & 79.5$\pm$1.6 & 79.8$\pm$1.6 & 63.6$\pm$1.7 & 63.4$\pm$1.3 & 40.3$\pm$0.7 & 38.6$\pm$1.3 & 73.6$\pm$3.3 & 74.0$\pm$4.0 \\
        1.0\%, 99.0\% & - & \textbf{95.3}$\pm$0.3 & - & \textbf{74.8}$\pm$0.4 & - & \textbf{42.8}$\pm$0.7 & - & \textbf{87.6}$\pm$1.6 \\
        5.0\%, not used & 90.2$\pm$0.8 & 90.1$\pm$0.8 & 70.6$\pm$0.8 & 70.8$\pm$0.6 & 41.9$\pm$0.6 & 42.6$\pm$0.9 & 84.3$\pm1.8$ & 84.4$\pm$1.4 \\
        5.0\%, 95.0\% & - & \textbf{96.2}$\pm$0.2 & - & \textbf{76.0}$\pm$0.4 & - & \textbf{44.1}$\pm$0.5 & - & \textbf{90.0}$\pm$0.7 \\
        \hline
        \end{tabular}
        \vspace{-5px}
    }
    \end{center}
    
    \label{tab:semi_res}
\end{table*}

\subsection{Semi-supervised Learning}
\label{sec:SSL}
We conducted experiments to evaluate the semi-supervised learning performance of our model on the aforementioned datasets.
As shown in Table \ref{tab:semi_res}, we verify the semi-supervised classification performance by adjusting the ratio of the labeled and unlabeled data.
Also, we compare our proposed model with the FAM.
Because the FAM model is based on  fully-supervised learning, we could not report the performance of the FAM on semi-supervised settings.
We calculated the average performance of 30 trials.

\subsection{Semi-supervised Continuous Learning}

In this experiment, we expanded our experiment to semi-supervised continuous learning.
We performed two experiments on NSynth dataset: (1) accuracy comparison between LPART with FAM by epochs, and (2) the uncertain sample rate and classification accuracies using the uncertainty measurement methods by epochs.
We also calculated the average performance of 10 trials.

\section{Results and Discussion}

\label{sec:result}

\subsection{Semi-supervised Learning}

The classification accuracies on the four datasets are summarized in Table \ref{tab:semi_res}, with different few-labeled data probability settings.
In all experimental setups, the best results were obtained from our model using both labeled and unlabeled data, which is much higher than the accuracy using the labeled data only.
The performance drops as the amount of labeled data decreases.
However, when the unlabeled data is used together, the performance gap is not significant, which means that unlabeled data plays an important role for classification when the number of labeled data is extremely limited.
In some datasets, such as CIFAR-10, the classification performance is not good.
This is because the extracted features were not well grouped by class (Figure \ref{fig:feature}-c).
When more appropriate feature extraction methods are used together, better results can be obtained.

\subsection{Semi-supervised Continuous Learning}
\begin{figure}[htb]
\begin{minipage}[b]{1.0\linewidth}
  \centering
  \centerline{\includegraphics[width=8.65cm]{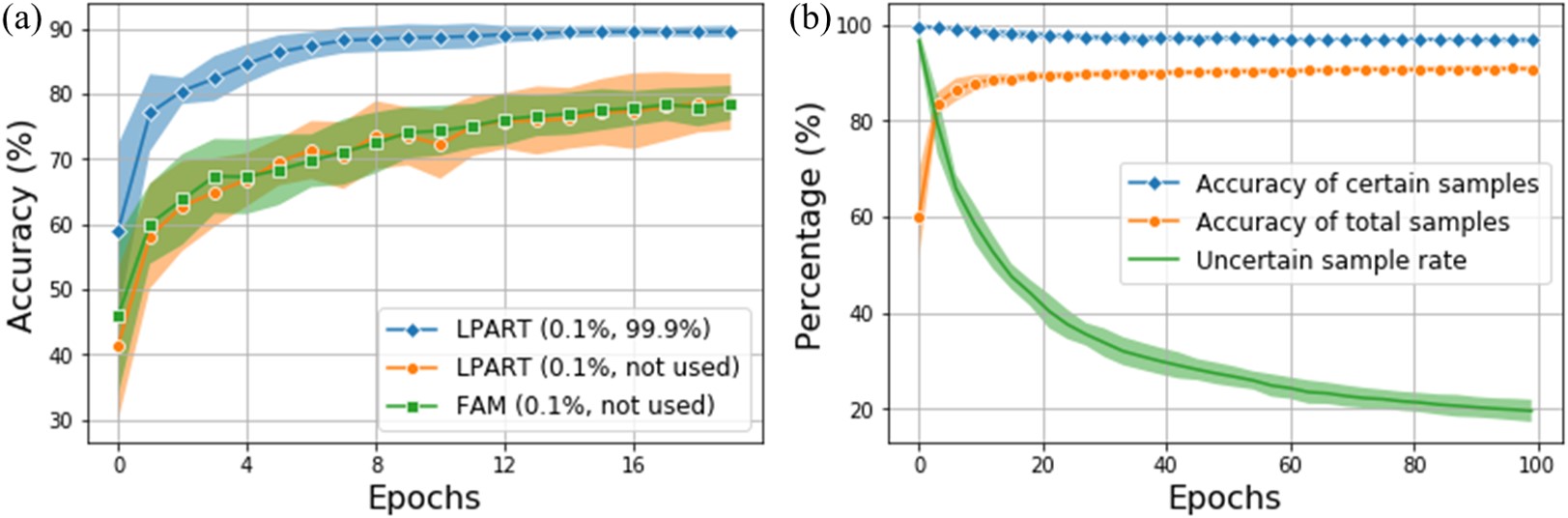}}
\end{minipage}

\vspace{-5px}
\caption{(a) Semi-supervised continuous learning result by epochs. (b) Uncertain sample rate and classification accuracies using the uncertainty measurement methods by epochs. The probability of the labeled sample was set to 0.1\%.}
\vspace{-5px}
\label{fig:res}
\end{figure}

The results of semi-supervised continuous learning using the NSynth dataset are shown in Figure \ref{fig:res}.
When unlabeled data is used together with the LPART, the classification performance per epoch rapidly increases to 90\% and converges without catastrophic forgetting (Figure \ref{fig:res}-a).
However, when only the labeled data was used, the classification performance increased slowly and shows a similar tendency to FAM's.
It confirms that the proposed model for semi-supervised continuous learning works properly.

We also filtered out the uncertain classification results using the uncertainty measurement methods described in Section \ref{sec:uncertainty}.
Thresholds for two uncertainty scores were set appropriately, and only the results with the scores below these thresholds remained (Figure \ref{fig:res}-b).
The reliable results always show high performance and the number of uncertain samples continue to decrease as the learning progresses.
This method is useful for applications where classification errors are fatal, and it allows selective use of reliable results in situations where labeled data is scarce and its collection is difficult.

\section{Conclusions}
\label{sec:conclusions}
In the present study, we proposed a novel approach for semi-supervised continuous learning based on ART networks.
We applied the label propagation mechanism to the ART network and evaluated it with various datasets and experimental settings to demonstrate its effectiveness.
Uncertainty measures can also be used to filter out unreliable classification results.
The limitation of this study is that a pre-trained feature extractor should be used, and the quality of the extracted feature can affect the overall performance.
In the future work, we will incorporate an end-to-end training of the entire system by applying a continuous learning method to the feature extractor.

\vfill\pagebreak

\bibliographystyle{IEEEbib}
\bibliography{strings,refs}

\end{document}